\documentclass{article} 
\usepackage{iclr2025_conference,times}

\usepackage{amsmath,amsfonts,bm}

\def\eqref#1{equation~\ref{#1}}

\def\1{\bm{1}}

\DeclareMathAlphabet{\mathsfit}{\encodingdefault}{\sfdefault}{m}{sl}
\SetMathAlphabet{\mathsfit}{bold}{\encodingdefault}{\sfdefault}{bx}{n}

\usepackage{url}
\usepackage{graphicx}
\usepackage{booktabs} 
\usepackage{enumitem}
\usepackage{xcolor}
\definecolor{cvprblue}{rgb}{0.21,0.49,0.74}
\usepackage[colorlinks=true,citecolor=cvprblue,]{hyperref}

\title{3DGS-Drag: Dragging Gaussians for \\Intuitive Point-Based 3D Editing}

\author{Jiahua Dong \quad Yu-Xiong Wang \\
University of Illinois Urbana-Champaign\\
\texttt{\{jiahuad2, yxw\}@illinois.edu} 
}

\newcommand{\rebuttal}[1]{%
     {{#1}}
}
\newcommand{\ours}{%
   3DGS-Drag }
\iclrfinalcopy 
\begin{document}

\maketitle
\vspace{-4mm}
\begin{figure}[h]
\centering
\includegraphics[width=1\textwidth]{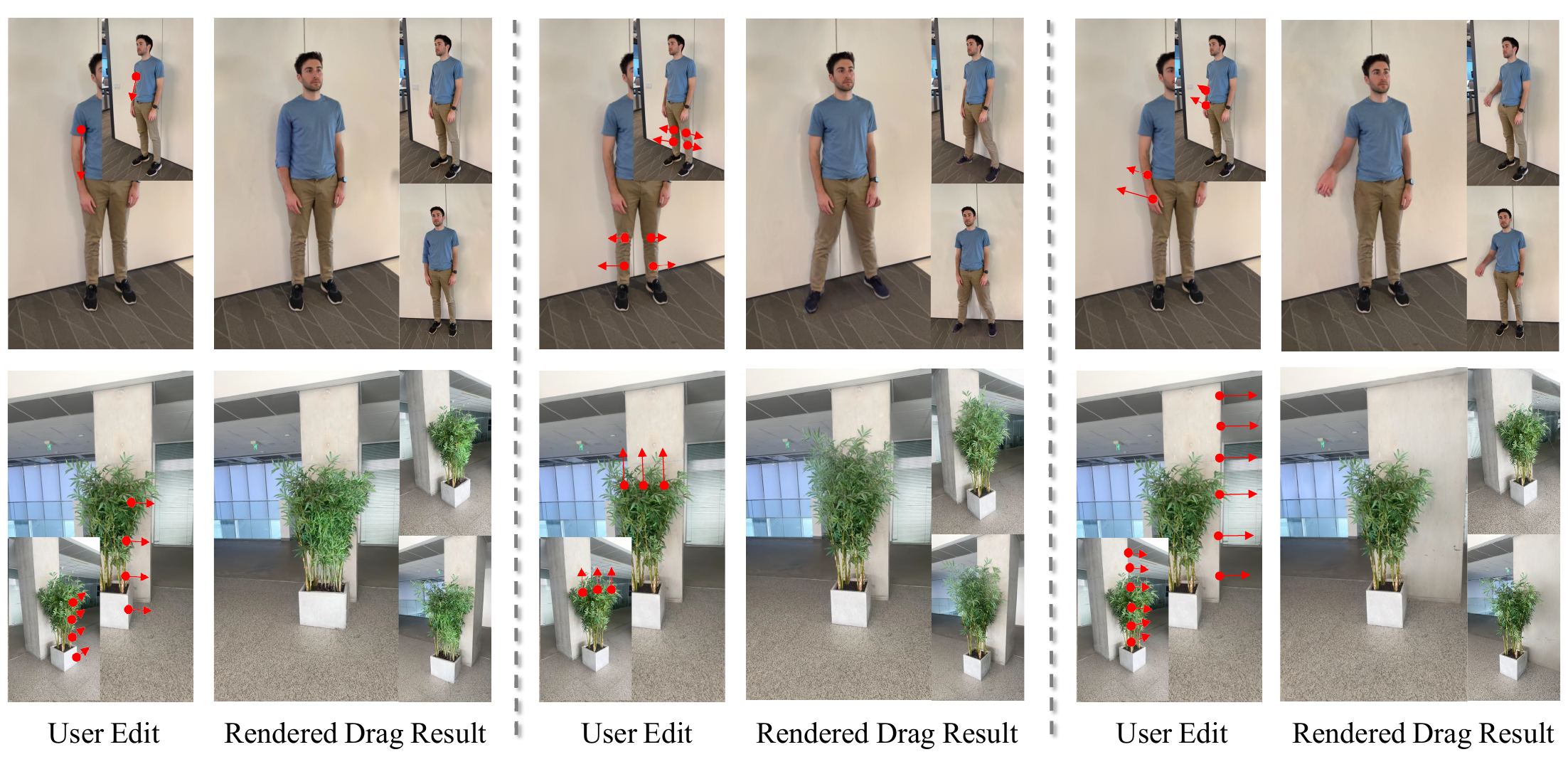}
\vspace{-8mm}
\caption{\textbf{Our proposed \ours framework enables high-quality 3D drag editing:} Users only need to input 3D handle points (\textcolor{red}{circle}) and target points (\textcolor{red}{triangle}). Our method precisely moves the handle points to match the target points while preserving the overall content and details.}

\label{fig:teaser}
\end{figure}

\begin{abstract}

The transformative potential of 3D content creation has been progressively unlocked through
advancements in generative models. Recently, intuitive drag editing with geometric changes has attracted significant attention in 2D editing yet remains challenging for 3D scenes. In this paper, we introduce \ours -- a point-based 3D editing framework that provides efficient, intuitive drag manipulation of real 3D scenes. Our approach bridges the gap between deformation-based and 2D-editing-based 3D editing methods, addressing their limitations to geometry-related content editing. We leverage two key innovations: \textit{deformation guidance} utilizing 3D Gaussian Splatting for consistent geometric modifications and \textit{diffusion guidance} for content correction and visual quality enhancement. A \textit{progressive} editing strategy further supports aggressive 3D drag edits. Our method enables a wide range of edits, including motion change, shape adjustment, inpainting, and content extension. Experimental results demonstrate the effectiveness of \ours in various scenes, achieving state-of-the-art performance in geometry-related 3D content editing. Notably, the editing is efficient, taking 10 to 20 minutes on a single RTX 4090 GPU. Our code is available at \href{https://github.com/Dongjiahua/3DGS-Drag}{https://github.com/Dongjiahua/3DGS-Drag}.
\end{abstract}

\section{Introduction}

\label{sec:intro}
Recent years have witnessed remarkable advancements in 3D scene representation techniques, such as Neural Radiance Fields (NeRF)~\citep{nerf} and 3D Gaussian Splatting (3DGS)~\citep{3dgs}. These methods have revolutionized the way we capture, represent, and synthesize 3D content, offering unprecedented levels of detail and realism. Inspired by their success and the blooming development of 2D generative models~\citep{sd}, recent works in 3D generation~\citep{dreamgaussian, dreamfusion} can now generate 3D content with high quality and efficiency. However, precise and intuitive editing of 3D scenes remains a challenge, particularly in contrast to the sophisticated editing capabilities available for 2D images. While 2D editing methods like DragGAN~\citep{draggan} offer point-based manipulation, extending such functionalities to 3D scenes presents substantial technical hurdles. 

Specifically, the underexplored capability behind is to achieve \textit{intuitive content editing} with \textit{geometric change}. The recent progress in 3D editing can be roughly grouped into two classes: deformation-based and 2D-editing-based. The deformation-based methods~\citep{scgs, physgaussian} primarily focus on motion editing, assuming strong geometry prior~\citep{physgaussian} or relying on video to learn motion pattern~\citep{scgs}. Besides the requirement for sufficient prior information, they naturally cannot intuitively edit unseen content. For 2D-editing-based methods, recent works~\citep{in2n, vica, consistentdreamer} have attempted to distill the editing ability from 2D diffusion models~\citep{ip2p} by editing the dataset of different view images with the 2D diffusion model. These approaches remain limited to appearance modifications and minor geometric adjustments, since larger 2D geometric edits fail to converge to 3D. The text guidance they used also sometimes causes incorrect edits, because the diffusion model fails to understand the text prompt. \textit{Bridging the geometric editing ability from deformation and the content editing ability from 2D-editing models has not yet been well studied}.

Motivated by these observations, we introduce \ours -- \emph{an intuitive 3D drag editing method for real scenes}. Extending the flexible editing format of DragGAN~\citep{draggan}, we take 3D handle points and target points as inputs, aiming for geometry-related 3D content editing.  Our key insight is to fully leverage two sources of guidance for 3D content editing, which explicitly regularize the edits from different views to be consistent and optimized toward the target 3D points. The first guidance is \emph{deformation guidance}. Benefiting from the explicit representation of 3D Gaussian Splitting~\citep{3dgs}, we propose a simple but effective deformation strategy without the requirement for prior information. With such a strategy, we directly deform the 3D Gaussians and leverage them as guidance for different views. Moreover, the deformation of the Gaussians facilitates optimization around the deformed space, simplifying the generation of detailed geometry. The second one is \emph{diffusion guidance}. Notably, since there is no prior information in our setting, the deformed Gaussians always have incorrect content and artifacts. We use the diffusion model to correct the content and improve the visual quality. This guidance is grounded in our observation that a fine-tuned diffusion model serves as a view-consistent editor for a 3D scene. Consequently, it achieves better consistency given previous deformation guidance.

To support more challenging 3D drag edits, we further propose a \textit{progressive} editing strategy. Specifically, we divide the drag operation into several intervals and proceed with editing step by step. The continuity of editing is guaranteed by a 3D relocation strategy. In the end, our experimental results demonstrate the effectiveness of our \ours in various scenes and editing. We resolve the challenges of a 3D drag operation and indicate an enhancement in multi-view consistency compared to prior techniques. 

Our major \textbf{contributions} can be summarized as follows: 1) We present a novel framework for editing 3D scenes, featuring a point-based drag editing approach. 2) We propose an effective method to bridge 3D deformation guidance and diffusion guidance for conducting geometry-related 3D content editing. 3) We further propose a progressive drag editing method to improve editing results. 4) Extensive evaluations show our method achieves state-of-the-art results in such setting, which implicitly includes motion change, shape adjustment, inpainting, and content extension.

\section{Related work}


\subsection{2D Image Editing}
Initially, the image generation methods rise from generative adversarial networks (GAN)~\citep{gan1,gan2}. Based on its latent representation, early works tried to modify the latent to adjust certain attributes or contents of the image~\citep{sgan1, sgan2, sgan3, sgan4}. However, due to the limited capability of the GAN model and the implicit representation of the latent code, it is hard to achieve high-quality and detailed edits. Recently, diffusion models have shown great potential for text-to-image tasks~\citep{sd}. Its feature map representation and the large-scale data empower lots of image editing methods~\citep{imagic,hy,sdedit,ip2p}. SDEdit~\citep{sdedit} performs a nosing and denoising procedure to keep the structural information and change the details. Instruct-Pix2Pix~\citep{ip2p} builds an instruction editing dataset and train the diffusion model to edit the image following the instruction. Compared with previous methods, Instruct-Pix2Pix shows better editing consistency.

Although text-based image editing can generate high-fidelity results, it cannot reach fine-grained editing.  DragGAN~\citep{draggan} proposed a point-based interactive editing method. The user inputs several handle points and target points; then, the latent will be optimized to move the handle points to the target. To improve the generality, DragDiffusion~\citep{dragdif} transfers this technique to diffusion models~\citep{sd}. Later, SDE-Drag~\citep{sde} and RegionDrag~\citep{regiondrag} further improve the performance. An inverse-forward process is necessary for these diffusion-based methods, making this operation time-consuming. In addition, there is no 3D consistency guaranteed in such 2D models, thus not available to be directly applied to 3D. 

In this paper, we adopt the 2D diffusion model to perform 3D-consistent view correction. Our editing not only generates intuitive new content but also removes potential 3D artifacts.

\subsection{2D-Editing-Based 3D Editing}
Previous to 3DGS~\citep{3dgs}, Neural Radiance Fields (NeRF)~\citep{nerf} is used as a common connector from 3D representation to 2D models. Early works on NeRF can only deal with color and shape adjustment~\citep{t1,t2,t3,t4,sine,arf,shop}. SNeRF~\citep{snerf} proposes to use an image stylization model, achieving high-quality stylization results. Later, NeRF-Art~\citep{nerfart} uses CLIP~\citep{clip} to distill the knowledge to NeRF. However, since the CLIP is not a generative model and is highly semantic-based, such an approach cannot get results with high fidelity. Instruct-NeRF2NeRF~\citep{in2n} proposes to use the Instruct-Pix2Pix model to Iteratively edit the dataset. They can edit various scenes with a broad range of instructions. ViCA-NeRF~\citep{vica} proposes to directly edit the dataset without fine-tuning NeRF. Specifically, they make multi-view consistent edits by utilizing the depth of information. DreamEditor~\citep{de} proposes to use a fine-tuned Dreambooth~\citep{dreambooth} to help with editing. ConsistentDreamer~\citep{consistentdreamer} further fine-tune a ControllNet to give more detailed edits. However, all these methods are limited by the 3D consistency from different views, thus only available to make subtle geometric changes. PDS~\citep{pds} propose a new distillation loss to help improve the result but suffer from degeneration of rendering quality and the ability for sufficient geometric editing. 

Inspired by the efficiency of 3DGS, recent approaches~\citep{cvpr_editor,gseditor, chen2024proedit} propose to migrate the success of NeRF editing to 3G Gaussians. However, they are mainly following the idea of Instrcut-NeRF2NeRF~\citep{in2n} by changing the 3D representation, thus having similar limitations. \rebuttal{Some approaches~\citep{drag1,draggaussian, apap, interactive3d} have attempted to extend the drag operation to 3D; however, they are limited to handling single objects.} In contrast, our approach leverages the explicit representation of 3DGS and focuses on real scenes. 

\subsection{Deformation-Based 3D Edting}
3D deformation is a challenging task since the target is to generate unseen motions. Traditional methods~\citep{meshd1, meshd2} apply certain Laplacian coordinates for mesh deformation. Recently, people have focused on deformation in 3D representations like NeRF and 3DGS. Specifically, \citet{cagenerf} proposes to build 3D cages as the motion prior to guide deformation. \citet{geoediting} reconstruct the mesh from NeRF and deform the mesh instead. NeuralEditor~\citep{neuraleditor} requires dense point cloud deformation as input and applies point-like NeRF structure for deformation. All these methods need strong geometry prior to editing, which is hard and inconvenient in practice. PhysGaussian~\citep{physgaussian} considers Gaussian ellipsoids as a Continuum and integrates physics. SC-GS~\citep{scgs} samples control points as a structure-representing graph to guide motion. However, the physics simulation and continuum assumption make PhysGaussian less flexible and limited to continuous scenes. SC-GS's control points are an approximation of dense points, thus also relying on sufficient capture of the object's geometry. It also takes dynamic scenes as input to build prior motion knowledge. 

The hard prior knowledge requirements or strict assumptions make these methods not suitable for large real scenes, where there is often only part-view information and complex layouts. In addition, they do not have the ability to create new parts. Rather than building a better deformation method, we propose a simpler deformation strategy for 3DGS to give rough deformations. Since the 2D generative models~\citep{sd} \emph{already have the sense of normal motions and contents}, we borrow such knowledge to provide more flexible 3D edits

\section{Method}\label{sec:method}
\begin{figure}[t]
\begin{center}
\includegraphics[width=1\textwidth]{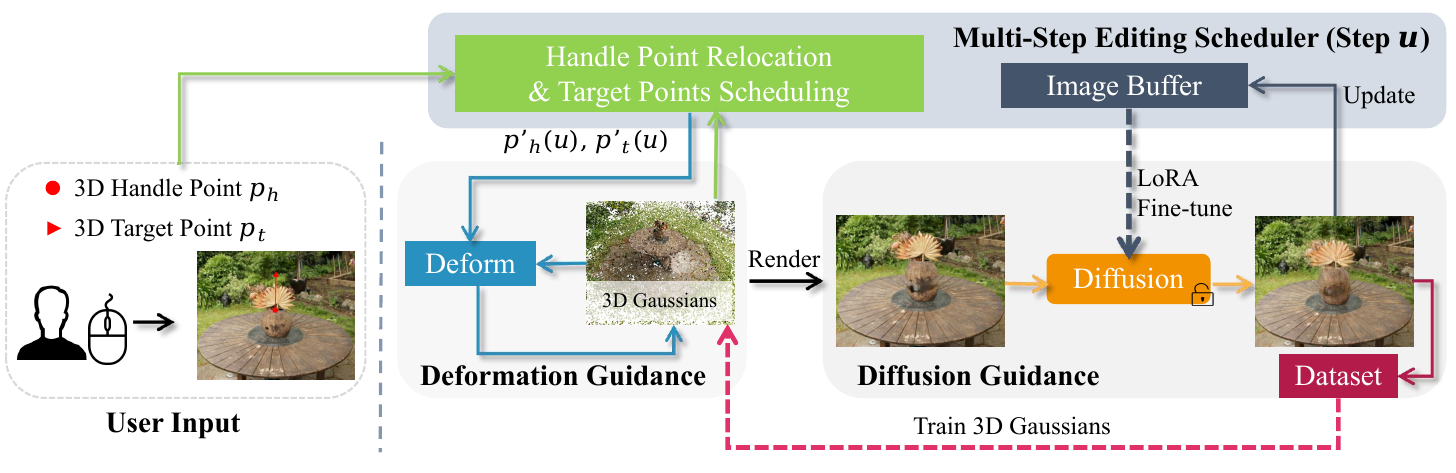}
\end{center} 
\vspace{-5mm}
\caption{\textbf{Overview of \ours:} Given a trained 3D Gaussian splatting model and the dataset, we use the multi-step editing scheduler to calculate the intermediate handle points $p'_h(i)$ and target points $p'_t(i)$ for step $i$. In each step, we first deform the 3D Gaussians using handle points and target points. Then, we render the image for each view and correct it with a diffusion model. The final corrected images will be used to train 3D Gaussians to improve quality. The diffusion model is fine-tuned with LoRA for more consistent edits.}
\vspace{-2mm}
\label{fig:pipe}
\end{figure}

\subsection{Preliminary}

\noindent{\textbf{3D Gaussian Splatting.}} 3D Gaussian splatting~\citep{3dgs} uses a collection of 3D Gaussians to represent 3D information, demonstrating effectiveness in object and scene reconstruction tasks. Each Gaussian is characterized by a center $\mu \in \mathbb{R}^3$, a scaling factor $s \in \mathbb{R}^3$, and a rotation quaternion $q \in \mathbb{R}^4$. The model also incorporates an opacity value $\alpha \in \mathbb{R}$ and a color feature $c \in \mathbb{R}^d$ for volumetric rendering, where $d$ indicates the degrees of freedom. The full set of parameters is denoted as $\Gamma$, where $\Gamma^i = \{\mu^i, s^i, q^i, \alpha^i, c^i\}$ represents the parameters for the $i$-th Gaussian.

\subsection{Framework Overview}

Our framework is illustrated in Figure~\ref{fig:pipe}. It takes a pretrained 3D Gaussian splatting model and several handle points along with their corresponding target points as input. Specifically, the handle points are denoted as $p_{h}^{n\times3}$, and the target points are denoted as $p_{t}^{n\times 3}$, where $n$ is the number of handle points. We aim to move the handle part to the target position while preserving similar content. Depending on the input points, this process may entail appearance and geometric changes, allowing more challenging edits with user-friendly inputs.

Different from the idea of 2D drag editing techniques~\citep{draggan, dragdif, regiondrag}, which either optimize or operate the inverse feature of a 2D image, we use \emph{deformation-based geometric guidance} and \emph{diffusion-based appearance guidance} for 3D editing. For a single step of drag operation, we first deform the 3D Gaussians with the provided handle and target points (Sec.~\ref{sec:deform}). Such deformation is conducted in a copy-and-paste manner to allow more editing flexibility. Due to the sparsity and long-distance challenge of the drag operation, the rendering result from the deformed Gaussians have poor visual quality and incorrect content. Thus, we propose to use diffusion-guided image correction on the rendered images (Sec.~\ref{sec:dif}), which efficiently corrects the contents and removes artifacts. To resolve editing with more aggressive changes, we propose a multi-step editing scheduler to progressively edit the scene (Sec.~\ref{sec:multistep}). As the whole process is divided into intervals, the user can stop at any intermediate step when achieving a satisfactory outcome.

\subsection{Deformation Guidance for Geometric Modification}\label{sec:deform}
As we aim to deform the 3D scenes to provide geometry guidance, we leverage 3DGS to benefit from its explicit representations and efficiency. The deformation involves two challenges in our task: (1) How to approximately deform the 3D Gaussians given sparse handle points and long-distance drag target, without structural modification to standard 3DGS; (2) How to avoid degeneration to direct deformation, allowing more flexibility to edits like moving, extending, and others. The proposed solution is described as follows. As a result, we achieve reliable deformation to 3DGS given limited point information.

\noindent\textbf{Drag Deformation.} 
The explicit representation of 3DGS enables efficient 3D deformation and adjustment. However, the real deformation function cannot be precisely computed, given only handle points and target points. Thus, we approximate it to give a rough geometry guidance. For the $i$th handle point $p_h^i$, we assign the Gaussians $P_h^i$ within a certain distance $\tau$ in 3D to this point. These Gaussians are considered to be deformed and guided by this handle point. The union of $\{P_h^i|i=1,2,...,n\}$ is denoted as $P_h = \bigcup_{i=1}^{n} P_h^i
$. 

Firstly, we calculate the translation and rotation for each handle point. For the translation, we simply calculate it as: $\Delta p_h^i=p_t^i - p_h^i$. For the rotation, it is not to further change the position of handle points but to represent the potential orientation change. Since the 3D Gaussians are also parameterized by rotation $q$, such a parameter is crucial to guide the Gaussian deformation. However, our handle points are just coordinates without information on the orientation. To approximate the rotation, we calculate its relative rotation with its \rebuttal{top-$K$ $(K=2)$} nearest handle points $\{p_h^k|k \in N_h^i\}$ where $N_h^i$ are the indices of \rebuttal{top-$K$} nearest handle points. Linear weight is used due to the sparsity of the points. Specifically, the weight is calculated as: 

\begin{equation}
w_h^{ik} = 1 - \frac{\left\| p_h^i - p_h^k \right\|_2^2}{\sum_{j \in N_h^i}\left\| p_h^i - p_h^j \right\|_2^2}.
\end{equation}

Then, we calculate the relative rotation quaternion $\Delta q_h^{ik}$ between $p_h^i$ and $p_h^k$ (Details in Sec.~\ref{sec:sup_deform}), and the quaternion $\Delta q_h^i$ of pair $(p_h^i, p_t^i)$ is calculated as $\Delta q_h^i = \sum_{k \in N_i}w_h^{ik}\Delta q_h^{ik}$.

After calculating each handle point's translation and rotation quaternion, we can interpolate the entire 3D Gaussians' deformation. Specifically for each Gaussian $\Gamma^i\in P_h$, the deformed Gaussian is interpolated from the transformation of \rebuttal{top-$K$ $(K=2)$} nearby handle points $\{p_h^j|j \in N_i\}$ where $N_i$ are the indices of \rebuttal{top-$K$} nearest handle points.  The deformed center $\mu_d^i$ and rotation quaternion $q_d^i$ are:
\begin{align}
w^{ik} &= 1-\frac{\left\| \mu^i - p_h^k \right\|_2^2}{\sum_{j \in N_i}\left\| \mu^i - p_h^j \right\|_2^2},\\
\mu_d^i &= \mu^i + \sum_{k \in N_i} \, w^{ik} \Delta p_h^k, \\
q_d^i &= \sum_{k \in N_i} (w^{ik}\Delta q_h^k)\otimes q^i,
\end{align}
where $\mu^i$ and $q^i$ are the original center and rotation quaternion. $\otimes$ is the quaternion production. When there is only one handle point, no quaternion change will be applied. We do not directly update the old Gaussians to the deformed Gaussians since this limits deformation and is not suitable for tasks like ``make his sleeves longer.'' Inspired by SDE-Drag~\citep{sde}, we use a copy-and-paste manner to place the deformed Gaussians and keep the old ones. To offer more flexibility for optimization, we adjust the opacity of the original Gaussians $P_h$ to a smaller value, allowing the 2D updates to determine whether to keep or remove the Gaussians.

\begin{figure}[t]
\begin{center}
\includegraphics[width=1\textwidth]{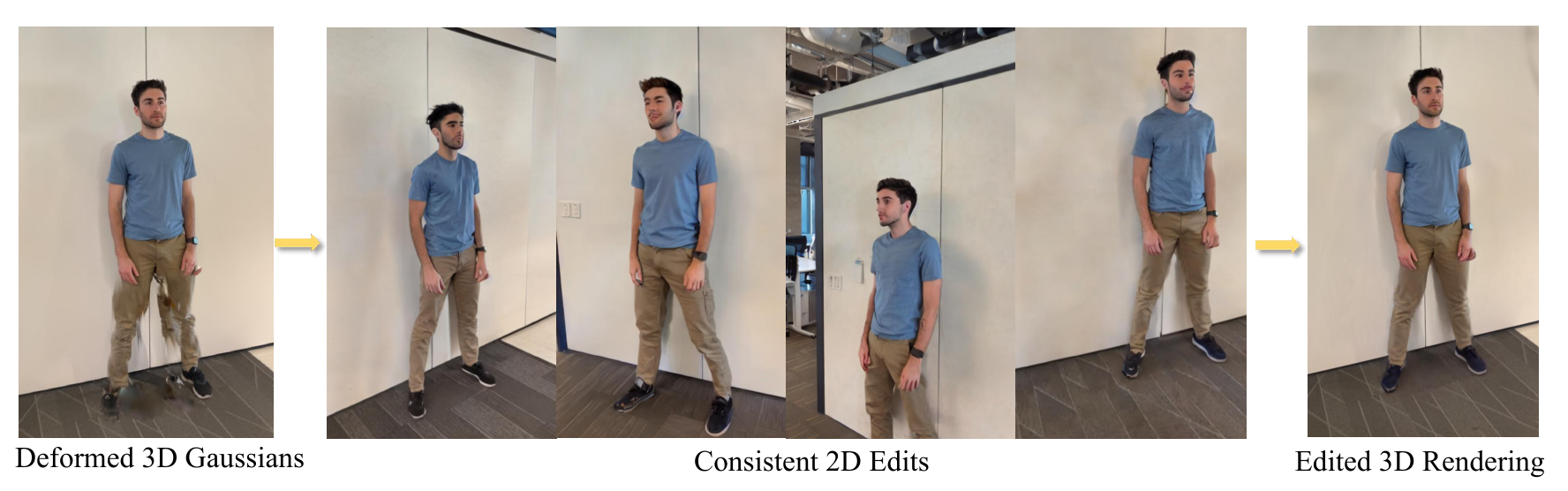}
\end{center} 
\vspace{-5mm}
\caption{\textbf{Multi-view consistent 2D edits:} With the deformed rendering as input, the fine-tuned diffusion model can perform multi-view consistent edits, and the artifacts and incorrect parts (shoes) are fixed.}
\vspace{-10pt}
\label{fig:consis}
\end{figure}
\noindent\textbf{Local Editing Mask.} 
Since drag operations mainly focus on a part of the entire scene, local editing is necessary to maintain the background information. Following Gaussian Editor~\citep{gseditor}, we assign a mask $M$ to the Gaussians of $P_h$, which are considered changeable. Different from Gaussian Editor, our work builds \textit{both 3D and 2D local editing masks} to work with more complex scenes and geometry edits. For the 3D mask, we inherit the mask from the original Gaussians when deforming new Gaussians or during the densification procedure. These Gaussians outside of the mask are not changed in the optimization. For the 2D mask, we render the mask for each view and round it to (0, 1) with a threshold, resulting in masks $\{m^v\}$ where $v$ denotes the $v$th view. Note that the mask rendering is after the deformation, so the original region and the target region will both be covered. The mask is further dilated to change the context of the nearby area.

\subsection{Diffusion Guidance for Appearance Correction}\label{sec:dif}
The direct deformation of Gaussians often creates notable artifacts and cannot generate correct semantic content. Inspired by recent successes in 3D editing~\citep{in2n}, we update the dataset to edit 3D scenes. However, integrating the concept of 2D dragging into a 3D context is non-trivial. Previous 2D drag methods often necessitate a time-consuming forward and backward process~\citep{dragdif, sde}. Moreover, during the training process, the inconsistent 2D edits from different views make the final result deviate from expectations and full of artifacts. To address these issues, we propose to use inverse-free 2D image editing that achieves \emph{stronger 3D consistency, efficiency, and quality}, relying on consistent renderings from the deformed 3D content. As shown in Figure~\ref{fig:consis}, our method generates multi-view consistent 2D edits. In detail, given the rendered image from the deformed 3D Gaussians, we introduce an \textit{Image2Image view correction} to obtain corrected 2D edits. To overcome the challenge of dataset editing with geometry change, we update the dataset in an \textit{annealed dataset editing} way.

\noindent\textbf{Image2Image View Correction.} Although the deformed Gaussian gives better 3D consistency, it cannot benefit from latent-based drag methods~\citep{draggan}. This is because the 3D consistency is ensured with newly rendered images. In contrast, latent-based methods heavily rely on operating the feature map of the same image. Inspired by the common approach for image editing~\citep{sdedit}, we add noise and then denoise it through the Dreambooth~\citep{dreambooth} model. By changing the image to a sketch level and denoising it, the diffusion model can partially understand and complete the deformed part. 

To mitigate the influence of randomness from the diffusion model, the Dreambooth model is fine-tuned on each scene with LoRA~\citep{lora}. We find that after fine-tuning, the diffusion model becomes a multi-view consistent editor. The experiment results in Figure~\ref{fig:consis} show that the diffusion model can successfully understand the deformed image and generate an image with the correct content even without the inverse process. However, such corrections still cannot fully converge in one update, requiring a better dataset editing strategy as follows.

\noindent\textbf{Annealed Dataset Editing.} Iterative dataset editing has been a common approach for 3D appearance editing ~\citep{in2n}. The idea is to progressively change the appearance of 3D and use the rendering to guide consistent 2D editing further. However, such a strategy does not work well with geometry-related edits because it is harder to converge given inconsistent geometry. In addition, long-term iterative updates also accumulate serious blurriness ~\citep{in2n}. To address this, we propose to update the dataset with limited $A$ times, and each time anneals the strength~\citep{sdedit} for Image2Image view correction. The annealing function is as follows:
\begin{equation}
S(a) = S_{\mathrm{init}} -\frac{a-1}{A}(S_{\mathrm{init}}-S_{\mathrm{final}}), \quad a=1,2,3,...,A,  
\end{equation}
where $S_{\mathrm{init}}$ and $S_{\mathrm{final}}$ are the initial strength and final strength respectively. $S(a)$ denotes the strength for the $a$th updates. Note that lower strength means that diffusion starts from later timesteps, resulting in finer detail correction. Our strategy performs editing in a coarse-to-fine manner. Each time, all the views are updated to prevent accumulated errors. 

\noindent\textbf{Loss Function.}
With the rendered image $I^v_r$ from 3D Gaussians, the corresponding edited image $I^v_e$ as the editing area's groundtruth, the original image $I^v_o$ as background groundtruth and mask $m^v$ for view $v$, our loss function for training 3D Gaussians is formulated as:
\begin{align}
    \mathcal{L} = \sum_{v=1}^V({\lambda_{1}\mathcal{L}_1(I^v_r,I^v_o) +\lambda_{\mathrm{ssim}} \mathcal{L}_{\mathrm{ssim}}(I^v_r,I^v_o))\odot(1-m^v)} + \lambda_{\mathrm{lpips}}\mathcal{L}_\mathrm{lpips}(I^v_r,I^v_e)\odot m^v),
\end{align}
where $\mathcal{L}_1$ and $\mathcal{L}_{\mathrm{ssim}}$ are to ensure local editing. $\mathcal{L}_\mathrm{lpips}$ is the LPIPS~\citep{lpips} loss function to correct the editing area. $\lambda_1$, $\lambda_{\mathrm{ssim}}$ and $\lambda_{\mathrm{lpips}}$ are the weighting coefficient for each loss.

\begin{figure}[t]
\centering
\includegraphics[width=1\textwidth]{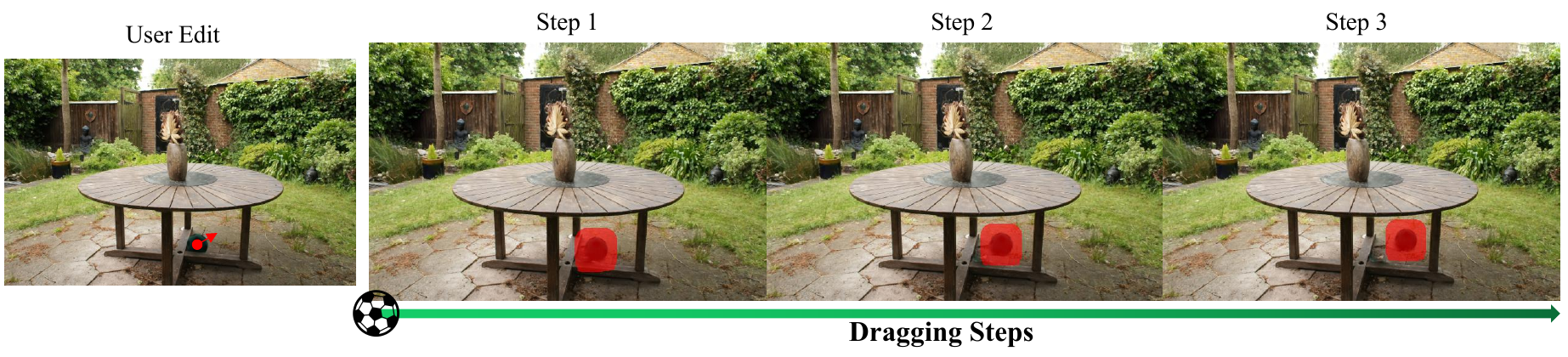}

\vspace{-4mm}
\caption{\textbf{Intermediate dragging steps and tracked mask:} Our method conducts progressive editing toward the target point. The dragged Gaussians are tracked to achieve aggressive edits. }
\vspace{-2mm}
\label{fig:drag_steps}
\end{figure}
\subsection{From One-Step to Multi-Step Drag Editing}\label{sec:multistep}
The previous sections introduce the one-step drag editing using our method. As the long-distance drag operation often requires more than one step to avoid corruption, we propose a multi-step editing scheduler to solve such problems. Specifically, we split the drag operation into $T$ intervals and set the progressive target points $\{p_t'(u)|u=1,2,..., T\}$. In each interval, we perform drag toward the corresponding target points: 
\begin{equation}
    p_t'(u) = p_h + \frac{u}{T}(p_t - p_h),
\end{equation}
However, the actual handle point position usually changes when training 3D Gaussians. We propose relocating the handle points at the end of every interval to make the next interval's deformation more precise. In addition, we further conduct history-aware diffusion fine-tuning to improve the ability for more aggressive editing.

\noindent\textbf{Handle Point Relocation.}
The handle point relocation is performed after each interval's training process. To keep track of the handle points, we use the Gaussians associated with each handle point. Specifically for handle point $p_h^i$,  we update it with the averaged position change of Gaussians $P_h^i$. \rebuttal{As shown in Figure~\ref{fig:drag_steps}, the dragged part can be successfully relocated.} Note that the assigned Gaussians $P_h^i$ are updated to newly deformed Gaussians during deformation and inherited from parents during the densification process of training. The local mask is updated as the union with the mask.

\noindent\textbf{History-Aware Diffusion Fine-Tuning.}
For long-distance drag operations, the edited 2D images can shift out of the diffusion model's domain since it is fine-tuned on the original images, resulting in degeneration back to the original images. We build an image buffer to fine-tune the diffusion model. The diffusion model will be fine-tuned with the image buffer every interval. Initially, the buffer only contains original images, and the newly edited result will be added to the buffer during intervals.

\begin{figure}[t]
\centering
\includegraphics[width=1\textwidth]{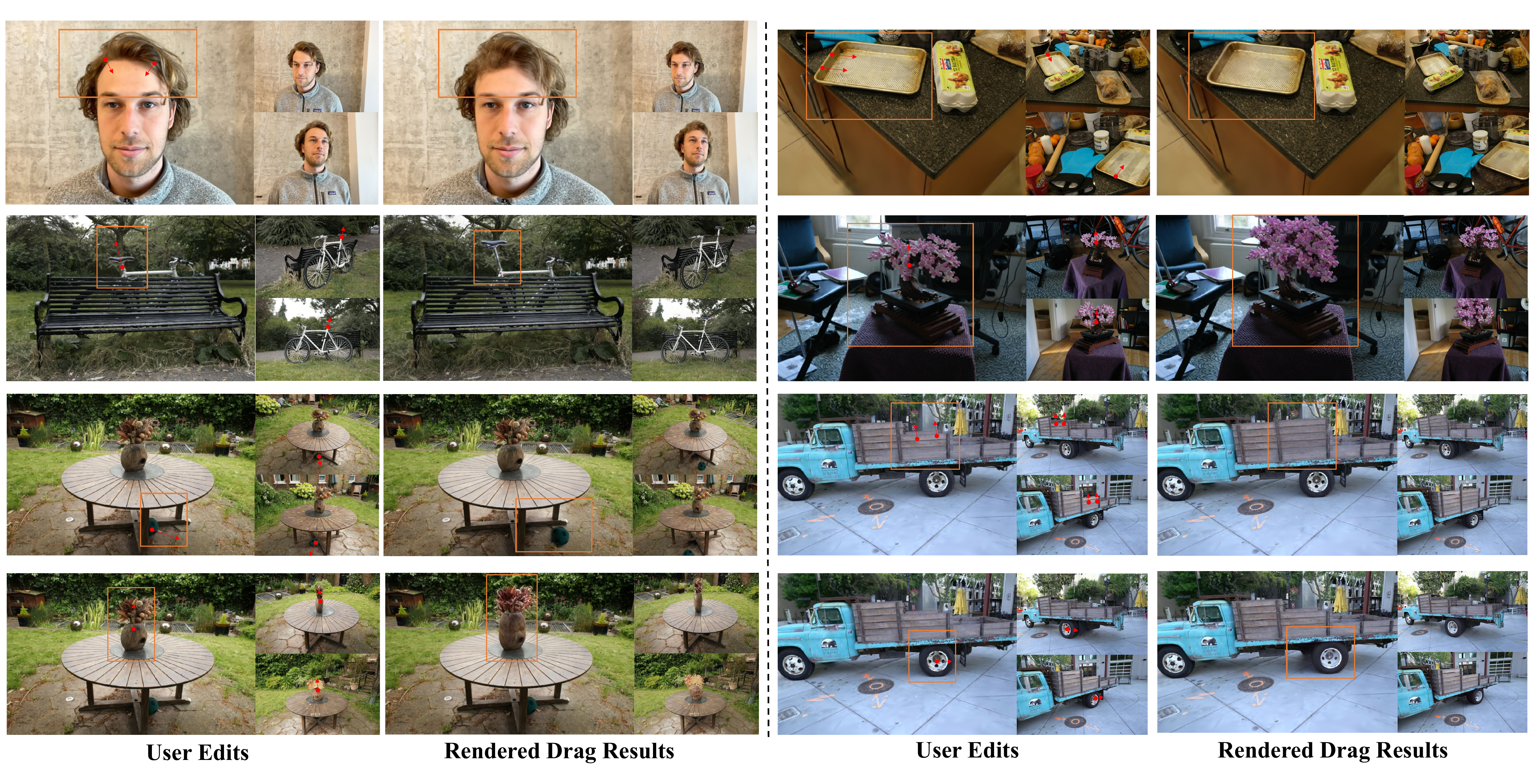}

\vspace{-5mm}
\caption{\textbf{Qualitative results in various scenes}: Our method can handle complex scenes and generate highly detailed results. With a simple drag input, \ours can identify the 3D context and perform edits like moving objects, inpainting the background, adjusting appearance, modifying object shape, and adjusting motion. \rebuttal{The orange bounding boxes highlight the modified regions.}}

\label{fig:real_res}
\end{figure}

\section{Experiment}

\subsection{Implementation Details}\label{sec:im_details}
\noindent\textbf{User Input.} Our user input is one or multiple handle points and corresponding target points. The input points are in 3D space. The user can specify the sphere radius of that handle point to adjust the editing scale. We automatically perform local editing by applying the mask rendered from assigned Gaussians. The mask is dilated to change the necessary context.

\noindent{\textbf{Drag Editing.}}
The pretrained 3D Gaussians are trained with original 3D Gaussian Splatting~\citep{3dgs}. During editing, 50 views are selected to enable efficient editing by default. Specifically, we choose the views with a larger visible area on the handle points' Gaussians, which is determined by the local editing mask on each view. We fine-tune the Dreambooth model~\citep{dreambooth} with LoRA~\citep{lora}. Initially, it is fine-tuned on the selected views. We use batch size 4 and train for 200 iterations. After each dragging step, we continue fine-tuning the diffusion model for 50 iterations with the updated image buffer in each interval. Each time, the newly edited image will be enqueued. The loss weight of $\lambda_1$, $\lambda_{\mathrm{ssim}}$ and $\lambda_{\mathrm{lpips}}$ are set to $8, 2, 1$ respectively. Note that the  $\lambda_1$ and $\lambda_{\mathrm{ssim}}$ are 10 times bigger than normal to ensure the background.

\noindent\textbf{Datasets.}\label{sec:dataset}
our experiments include edits on eight scenes, using the published datasets from Instruct-NeRF2NeRF~\cite{in2n}, PDS~\cite{pds}, Mip-NeRF360~\cite{mip}, and Tank and Temple~\cite{tank}.

\subsection{Qualitative Evaluation}
\label{sec:qual}
\noindent\textbf{Editing Results in Various Scenes.} We show editing results from different views in Figure~\ref{fig:teaser} and Figure~\ref{fig:real_res}. Since the handle and target points are in 3D, we plot them in 2D for illustration. Each drag is represented by a red arrow where the start is the handle point, and the end is the target point. In the standing-person scene in Figure~\ref{fig:teaser}, when we raise one hand, this is very challenging since the arm is only observed partially, and the part under the arm is unknown. Our method also shows the ability to generate new poses and fix the texture on the pants below the arm. We are also able to change the leg motion and extend the sleeves. When dealing with more complex scenes, such as the bamboo scene in  Figure~\ref{fig:teaser}, \ours can understand the texture of the plant and extend it to be taller or wider. We can also easily change part of the background, like the wall. When the drag operation is to move the football, we can separate this object from the background and inpaint the texture at the original position instead of an empty region. In short, our drag operation can understand different operations in front-view or 360-degree scenes, such as moving objects and extending objects, demonstrating the ability to identify the 3D context.

\begin{figure}[t]
\begin{center}
\includegraphics[width=1\textwidth]{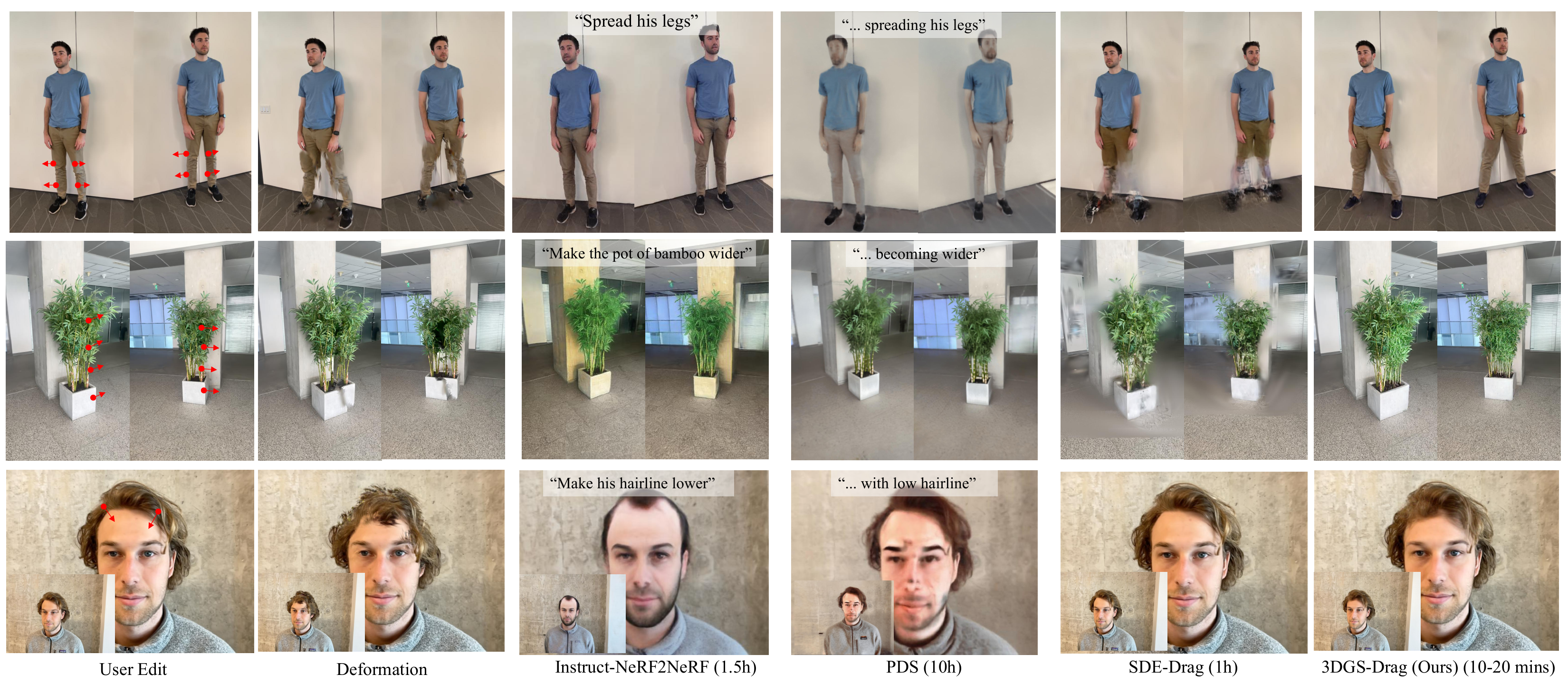}
\end{center} 
\vspace{-5mm}
\caption{\textbf{Baseline comparisons:} Compared with baselines, \ours achieves high-quality, fine-grained editing by correctly modifying different parts and in terms of efficiency. Specifically, Instruct-NeRF2NeRF~\citep{in2n} and PDS~\citep{pds} cannot correctly edit. Deformation results in incomplete edits, and SDE-Drag~\citep{sde} sometimes fails to make changes.}
\vspace{-10pt}
\label{fig:ablation}
\end{figure}
\noindent\textbf{Baseline Comparison.}
\rebuttal{Since there is no directly comparable work on intuitive 3D drag operation in real scenes,} we extend and re-purpose representative baselines. The results are shown in Figure~\ref{fig:ablation}. Specifically, the comparison with baselines is listed as follows:
\begin{itemize}[noitemsep, topsep=0pt, leftmargin=*]
    \item \textit{Instruct-NeRF2NeRF}~\citep{in2n}: We manually create text descriptions for drag operations in this baseline. Then, we use Instruct-NeRF2NeRF to edit the scene. The model fails to give edits for the `person' scene. For the more complex `garden' scene, Instruct-NeRF2NeRF just blurs the rendering. This demonstrates its insufficient ability to perform geometric modification.
    \item \textit{Deformation}: We use our deformation to represent the previous deformation-based approaches since we have different input settings. Notably, the geometry is moved, which results in a lot of incorrect content and artifacts.
    \item \textit{PDS}: PDS~\citep{pds}  claims to be able to change the geometry, but this method struggles in all three editing scenarios. In addition, PDS tends to create noisy and blurred editing results compared with others.
    \item  \textit{SDE-Drag}: One alternative solution is to simply use the 2D drag method on each view. Here we choose SDE-Drag~\citep{sde} in comparison. However, such a strategy cannot reach consistent edits, resulting in flawed results or failure cases in editing.
\end{itemize}
Compared with these baselines, our methods achieve significantly better editing results, with better details and correct content.  Remarkably, for the ``lower his hairline'' text prompt, both Instruct-NeRF2NeRF and PDS misunderstand the text and make the hairline higher, which further emphasizes the importance of intuitive 3D editing.

\begin{figure}[t]
\begin{center}
\includegraphics[width=1\textwidth]{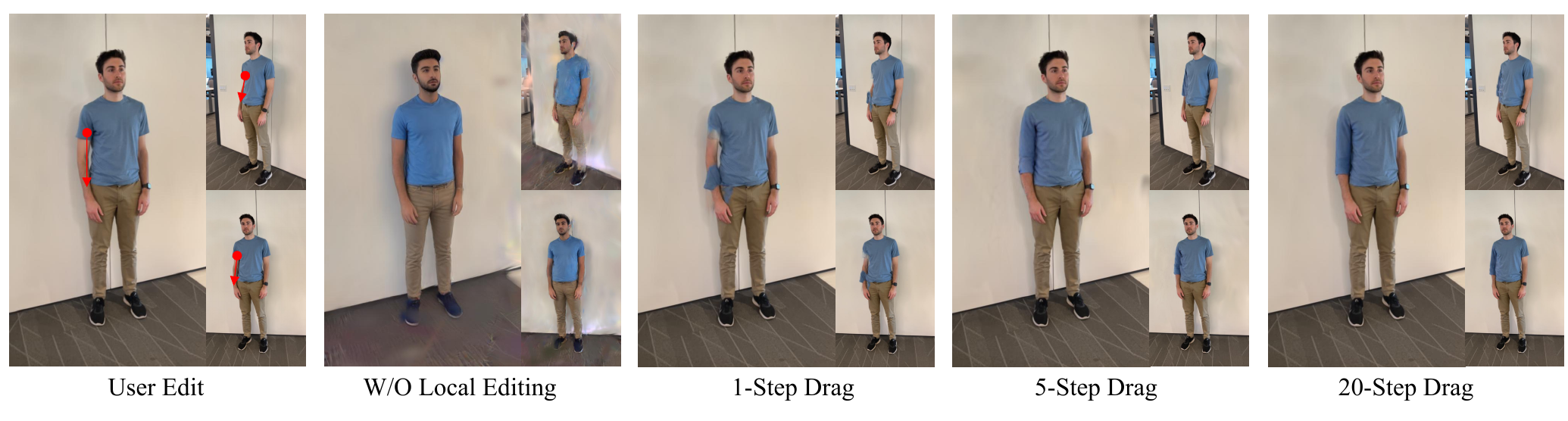}
\end{center} 
\vspace{-5mm}
\caption{\textbf{Ablation study on the local mask and drag steps:} Without the local mask, the scene will be blurred, resulting in failed edits. Using very few steps makes it hard to achieve aggressive edits. More steps will slightly improve the performance.}
\vspace{-4mm}
\label{fig:ad_abla}
\end{figure}
\noindent\textbf{Ablation Study} 
The diffusion guidance's effectiveness is validated when compared with the deformation approach (Figure~\ref{fig:ablation}). Here, We further ablate the local mask and multi-step strategies in Figure~\ref{fig:ad_abla}. (1) When local editing is not applied, the entire scene is blurred, and the edits fail. This is due to the optimization issue: inconsistent edits will create large floats in 3D Gaussians. (2) For drag steps, we compare three different drag steps from $[1,5,20]$, finding that more or fewer steps lead to different insights. When using a single step, the deformed Gaussians cannot give enough guidance to the diffusion model, resulting in broken edits. Thus, one-step drag editing usually meets challenges when we have more aggressive edits. When applying more steps (20 steps), the editing quality is slightly improved. This illustrates that \ours is robust when updated more times. However, since more steps will slow the execution, choosing an appropriate number of steps is better.

\subsection{Quantitative Evaluation} 
\begin{figure}[h]
\centering
\vspace{-3mm}
\includegraphics[width=.9\textwidth]{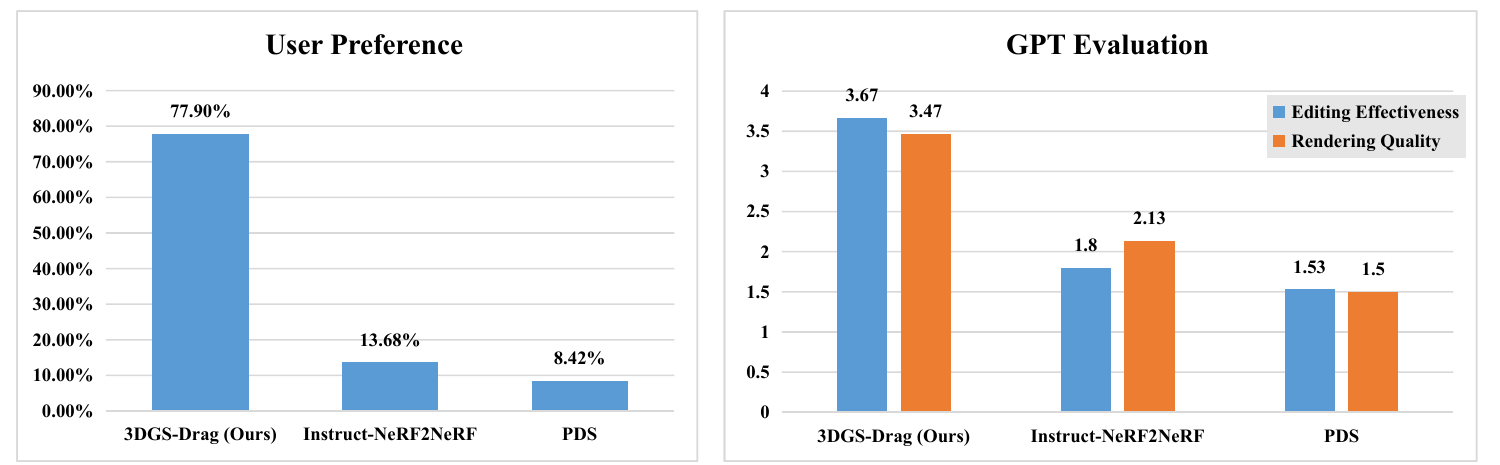}

\vspace{-3mm}
\caption{\textbf{Quantitative evaluation:} We conduct both user study and GPT evaluation on the editing results. Compared with Instruct-NeRF2NeRF~\citep{in2n} and PDS~\citep{pds}, \ours performs significantly better.}

\label{fig:quant}
\end{figure}

Quantitatively evaluating 3D editing results is often challenging since there lacks ground truth. Here, we use two metrics for evaluation: user preference and GPT score, shown in Figure~\ref{fig:quant}. For user preference, we conducted a user study across 19 subjects and collected their preference for each edit. For the GPT score, since GPT with vision has been proven to be a human-aligned evaluator~\citep{gptv}, we use gpt4-o to evaluate each editing, rating in 5 levels. Specifically, we measure (1) whether the content is correctly edited and (2) the rendered image quality for each method. Our method achieves the best results on all these metrics.
\subsection{Discussion}
\noindent\textbf{Limitations.}
Similar to previous diffusion-based 3D editing methods~\citep{gseditor, in2n}, our approach relies on the diffusion model to provide accurate guidance. Thus, our method may yield suboptimal results when the target object is too small within the field of view or when the scene exhibits considerable size and complexity. We also cannot deal with drag operations that are too aggressive. In such cases, the object may be relocated to areas with restricted visibility, which is out of vision for most views. 

\noindent\textbf{Running Time.}
When using 50 views for editing, our method needs 15 minutes. Specifically, about 2 minutes are needed for initial diffusion model fine-tuning, and 13 minutes are needed for the rest of the editing process. In comparison, Instruct-NeRF2NeRF~\citep{in2n} needs one hour. The running time is tested on a single RTX 4090 GPU.
\section{Conclusion}
In this paper, we introduced \ours, an intuitive drag editing approach for 3D scenes. In contrast to previous work~\citep{in2n, vica, nerfart}, which mainly focuses on appearance, we address the challenge of geometry-related content editing. Empirical experiments show that our method can achieve highly detailed edits across various scenes. Such an advantage stems primarily from our two key contributions: the copy-and-paste Gaussian deformation and the diffusion correction. We showcase that our method enables previously challenging edits, paving the way for exploring new possibilities in 3D editing. 

\newpage
\subsubsection*{Acknowledgments}
This work was supported in part by NSF Grant 2106825, NIFA Award 2020-67021-32799, the Toyota Research Institute, the IBM-Illinois Discovery Accelerator Institute, the Amazon-Illinois Center on AI for Interactive Conversational Experiences, Snap Inc., and the Jump ARCHES endowment through the Health Care Engineering Systems Center at Illinois and the OSF Foundation. This work used computational resources, including the NCSA Delta and DeltaAI supercomputers through allocations CIS220014 and CIS230012 from the Advanced Cyberinfrastructure Coordination Ecosystem: Services \& Support (ACCESS) program, as well as the TACC Frontera supercomputer, Amazon Web Services (AWS), and OpenAI API through the National Artificial Intelligence Research Resource (NAIRR) Pilot.

\subsubsection*{Reproducibility Statement }
Our code is released at \href{https://github.com/Dongjiahua/3DGS-Drag}{https://github.com/Dongjiahua/3DGS-Drag}. For the implementation details, we have covered our mathematical details in Sec.~\ref{sec:deform} and training details in Sec.~\ref{sec:im_details}. The framework architecture is fully introduced in Sec.~\ref{sec:method}. All the datasets we used are publicly available, as explained in Sec.~\ref{sec:dataset}.

\bibliography{iclr2025_conference}

@inproceedings{interactive3d,
  title={Interactive3D: Create What You Want by Interactive 3D Generation},
  author={Dong, Shaocong and Ding, Lihe and Huang, Zhanpeng and Wang, Zibin and Xue, Tianfan and Xu, Dan},
  booktitle={CVPR},
  pages={4999--5008},
  year={2024}
}

@inproceedings{apap,
  title={As-Plausible-As-Possible: Plausibility-Aware Mesh Deformation Using 2D Diffusion Priors},
  author={Yoo, Seungwoo and Kim, Kunho and Kim, Vladimir G and Sung, Minhyuk},
  booktitle={CVPR},
  pages={4315--4324},
  year={2024}
}

@inproceedings{regiondrag,
  title={RegionDrag: Fast Region-Based Image Editing with Diffusion Models},
  author={Lu, Jingyi and Li, Xinghui and Han, Kai},
  booktitle={ECCV},
  year={2024}
}

@article{chen2024proedit,
  title={ProEdit: Simple Progression is All You Need for High-Quality 3D Scene Editing},
  author={Chen, Jun-Kun and Wang, Yu-Xiong},
  journal={NeurIPS},
  year={2024}
}

@InProceedings{consistentdreamer,
    author    = {Chen, Jun-Kun and Bul\`o, Samuel Rota and M\"uller, Norman and Porzi, Lorenzo and Kontschieder, Peter and Wang, Yu-Xiong},
    title     = {ConsistDreamer: 3D-Consistent 2D Diffusion for High-Fidelity Scene Editing},
    booktitle = {CVPR},
    year      = {2024},
}

@inproceedings{lpips,
  title={The unreasonable effectiveness of deep features as a perceptual metric},
  author={Zhang, Richard and Isola, Phillip and Efros, Alexei A and Shechtman, Eli and Wang, Oliver},
  booktitle={CVPR},
  year={2018}
}

@inproceedings{physgaussian,
  title={Physgaussian: Physics-integrated 3d gaussians for generative dynamics},
  author={Xie, Tianyi and Zong, Zeshun and Qiu, Yuxing and Li, Xuan and Feng, Yutao and Yang, Yin and Jiang, Chenfanfu},
  booktitle={CVPR},
  pages={4389--4398},
  year={2024}
}

@inproceedings{scgs,
  title={Sc-gs: Sparse-controlled gaussian splatting for editable dynamic scenes},
  author={Huang, Yi-Hua and Sun, Yang-Tian and Yang, Ziyi and Lyu, Xiaoyang and Cao, Yan-Pei and Qi, Xiaojuan},
  booktitle={CVPR},
  year={2024}
}

@inproceedings{sde,
  title={The Blessing of Randomness: SDE Beats ODE in General Diffusion-based Image Editing},
  author={Nie, Shen and Guo, Hanzhong Allan and Lu, Cheng and Zhou, Yuhao and Zheng, Chenyu and Li, Chongxuan},
  booktitle={ICLR},
  year={2024}
}

@inproceedings{dragdif,
  title={DragDiffusion: Harnessing Diffusion Models for Interactive Point-based Image Editing},
  author={Shi, Yujun and Xue, Chuhui and Pan, Jiachun and Zhang, Wenqing and Tan, Vincent YF and Bai, Song},
  booktitle={CVPR},
  year={2024}
}

@inproceedings{draggan,
  title={Drag your gan: Interactive point-based manipulation on the generative image manifold},
  author={Pan, Xingang and Tewari, Ayush and Leimk{\"u}hler, Thomas and Liu, Lingjie and Meka, Abhimitra and Theobalt, Christian},
  booktitle={SIGGRAPH},

  year={2023}
}

@inproceedings{dreamfusion,
  title={Dreamfusion: Text-to-3d using 2d diffusion},
  author={Poole, Ben and Jain, Ajay and Barron, Jonathan T and Mildenhall, Ben},
  booktitle={ICLR},
  year={2023}
}

@inproceedings{in2n,
  title={Instruct-nerf2nerf: Editing 3d scenes with instructions},
  author={Haque, Ayaan and Tancik, Matthew and Efros, Alexei A and Holynski, Aleksander and Kanazawa, Angjoo},
  booktitle={ICCV},
  year={2023}
}

@inproceedings{dreamgaussian,
  title={Dreamgaussian: Generative gaussian splatting for efficient 3d content creation},
  author={Tang, Jiaxiang and Ren, Jiawei and Zhou, Hang and Liu, Ziwei and Zeng, Gang},
  booktitle={ICLR},
  year={2024}
}

@inproceedings{gseditor,
  title={Gaussianeditor: Swift and controllable 3d editing with gaussian splatting},
  author={Chen, Yiwen and Chen, Zilong and Zhang, Chi and Wang, Feng and Yang, Xiaofeng and Wang, Yikai and Cai, Zhongang and Yang, Lei and Liu, Huaping and Lin, Guosheng},
  booktitle={CVPR},
  year={2024}
}

@InProceedings{neuraleditor,
    author    = {Chen, Jun-Kun and Lyu, Jipeng and Wang, Yu-Xiong},
    title     = {NeuralEditor: Editing Neural Radiance Fields via Manipulating Point Clouds},
    booktitle = {CVPR},
    year={2023}
}

@InProceedings{lora,
  title={Lora: Low-rank adaptation of large language models},
  author={Hu, Edward J and Shen, Yelong and Wallis, Phillip and Allen-Zhu, Zeyuan and Li, Yuanzhi and Wang, Shean and Wang, Lu and Chen, Weizhu},
  booktitle={ICLR},
  year={2022}
}

@Article{3dgs,
      author       = {Kerbl, Bernhard and Kopanas, Georgios and Leimk{\"u}hler, Thomas and Drettakis, George},
      title        = {3D Gaussian Splatting for Real-Time Radiance Field Rendering},
      journal      = {ACM Transactions on Graphics},
      year         = {2023},
}

@InProceedings{sd,
      title={High-Resolution Image Synthesis with Latent Diffusion Models}, 
      author={Robin Rombach and Andreas Blattmann and Dominik Lorenz and Patrick Esser and Björn Ommer},
      year={2022},
      booktitle={CVPR},
}

@InProceedings{draggaussian,
  title={DragGaussian: Enabling Drag-style Manipulation on 3D Gaussian Representation},
  author={Shen, Sitian and Xu, Jing and Yuan, Yuheng and Yang, Xingyi and Shen, Qiuhong and Wang, Xinchao},
  booktitle={CVPR},
  year={2024}
}

@inproceedings{gptv,
  title={Gpt-4v (ision) is a human-aligned evaluator for text-to-3d generation},
  author={Wu, Tong and Yang, Guandao and Li, Zhibing and Zhang, Kai and Liu, Ziwei and Guibas, Leonidas and Lin, Dahua and Wetzstein, Gordon},
  booktitle={CVPR},
  year={2024}
}

@article{nerf,
  title={Nerf: Representing scenes as neural radiance fields for view synthesis},
  author={Mildenhall, Ben and Srinivasan, Pratul P and Tancik, Matthew and Barron, Jonathan T and Ramamoorthi, Ravi and Ng, Ren},
  journal={Communications of the ACM},
  year={2021},
}

@inproceedings{meshd1,
  title={As-rigid-as-possible surface modeling},
  author={Olga Sorkine-Hornung and Marc Alexa},
  booktitle={Eurographics Symposium on Geometry Processing},
  year={2007},
}

@inproceedings{geoediting,
  title={Nerf-editing: geometry editing of neural radiance fields},
  author={Yuan, Yu-Jie and Sun, Yang-Tian and Lai, Yu-Kun and Ma, Yuewen and Jia, Rongfei and Gao, Lin},
  booktitle={CVPR},
  year={2022}
}

@inproceedings{cagenerf,
  title={Deforming radiance fields with cages},
  author={Xu, Tianhan and Harada, Tatsuya},
  booktitle={ECCV},
  year={2022},
}

@article{meshd2,
  title={Laplacian mesh processing},
  author={Sorkine, Olga},
  journal={Eurographics (State of the Art Reports)},
  year={2005}
}

@inproceedings{ip2p,
  title={Instructpix2pix: Learning to follow image editing instructions},
  author={Brooks, Tim and Holynski, Aleksander and Efros, Alexei A},
  booktitle={CVPR},
  year={2023}
}

@inproceedings{vica,
      author = {Dong, Jiahua and Wang, Yu-Xiong},
      title = {ViCA-NeRF: View-Consistency-Aware 3D Editing of Neural Radiance Fields},
      booktitle = {NeurIPS},
      year = {2023},
     }

@ARTICLE{nerfart,
  author={Wang, Can and Jiang, Ruixiang and Chai, Menglei and He, Mingming and Chen, Dongdong and Liao, Jing},
  journal={IEEE Transactions on Visualization and Computer Graphics}, 
  title={NeRF-Art: Text-Driven Neural Radiance Fields Stylization}, 
  year={2023},
}

@inproceedings{cvpr_editor,
  author = {Fang, Jiemin and Wang, Junjie and Zhang, Xiaopeng and Xie, Lingxi and Tian, Qi},
  title = {GaussianEditor: Editing 3D Gaussians Delicately with Text Instructions},
  year = {2024},
  booktitle = {CVPR}
}

@inproceedings{gan1,
  title={Generative adversarial nets},
  author={Goodfellow, Ian and Pouget-Abadie, Jean and Mirza, Mehdi and Xu, Bing and Warde-Farley, David and Ozair, Sherjil and Courville, Aaron and Bengio, Yoshua},
  booktitle={NeurIPS},
year={2014}
}

@inproceedings{gan2,
  title={A style-based generator architecture for generative adversarial networks},
  author={Karras, Tero and Laine, Samuli and Aila, Timo},
  booktitle={CVPR},
  year={2019}
}

@article{sgan1,
  title={Styleflow: Attribute-conditioned exploration of stylegan-generated images using conditional continuous normalizing flows},
  author={Abdal, Rameen and Zhu, Peihao and Mitra, Niloy J and Wonka, Peter},
  journal={ACM Transactions on Graphics},
  year={2021},
}

@inproceedings{sgan2,
  title={User-Controllable Latent Transformer for StyleGAN Image Layout Editing},
  author={Endo, Yuki},
  booktitle={Computer Graphics Forum},
  year={2022},

}

@inproceedings{sgan3,
  title={Ganspace: Discovering interpretable gan controls},
  author={H{\"a}rk{\"o}nen, Erik and Hertzmann, Aaron and Lehtinen, Jaakko and Paris, Sylvain},
  booktitle={NeurIPS},
  year={2020}
}

@inproceedings{sgan4,
  title={Freestylegan: Free-view editable portrait rendering with the camera manifold},
  author={Leimk{\"u}hler, Thomas and Drettakis, George},
  booktitle={SIGGRAPH Asia},
  year={2021}
}

@inproceedings{imagic,
      title={{Imagic}: Text-Based Real Image Editing with Diffusion Models},
      author={Kawar, Bahjat and Zada, Shiran and Lang, Oran and Tov, Omer and Chang, Huiwen and Dekel, Tali and Mosseri, Inbar and Irani, Michal},
      booktitle={CVPR},
      year={2023}
}

@inproceedings{sdedit,
      title={{SDE}dit: Guided Image Synthesis and Editing with Stochastic Differential Equations},
      author={Chenlin Meng and Yutong He and Yang Song and Jiaming Song and Jiajun Wu and Jun-Yan Zhu and Stefano Ermon},
      booktitle={ICLR},
      year={2022},
}

@inproceedings{hy,
  title={{Hierarchical text-conditional image generation with {CLIP} latents}},
  author={Ramesh, Aditya and Dhariwal, Prafulla and Nichol, Alex and Chu, Casey and Chen, Mark},
  booktitle={arXiv preprint arXiv:2204.06125},
  year={2022}
}

@inproceedings{clip,
  title={{Learning transferable visual models from natural language supervision}},
  author={Radford, Alec and Kim, Jong Wook and Hallacy, Chris and Ramesh, Aditya and Goh, Gabriel and Agarwal, Sandhini and Sastry, Girish and Askell, Amanda and Mishkin, Pamela and Clark, Jack and Krueger, Gretchen and
Sutskever,Ilya},
  booktitle={ICML},
  year={2021}
}

@inproceedings{de,
  title={Dreameditor: Text-driven 3d scene editing with neural fields},
  author={Zhuang, Jingyu and Wang, Chen and Lin, Liang and Liu, Lingjie and Li, Guanbin},
  booktitle={SIGGRAPH Asia},
  year={2023}
}

@inproceedings{dreambooth,
  title={Dreambooth: Fine tuning text-to-image diffusion models for subject-driven generation},
  author={Ruiz, Nataniel and Li, Yuanzhen and Jampani, Varun and Pritch, Yael and Rubinstein, Michael and Aberman, Kfir},
  booktitle={CVPR},
  year={2023}
}

@inproceedings{t1,
  title={{Stylizing {3D} scene via implicit representation and hypernetwork}},
  author={Chiang, Pei-Ze and Tsai, Meng-Shiun and Tseng, Hung-Yu and Lai, Wei-Sheng and Chiu, Wei-Chen},
  booktitle={WACV},
  year={2022}
}

@inproceedings{t2,
  title={Learning to stylize novel views},
  author={Huang, Hsin-Ping and Tseng, Hung-Yu and Saini, Saurabh and Singh, Maneesh and Yang, Ming-Hsuan},
  booktitle={ICCV},
  year={2021}
}

@inproceedings{t3,
  title={{StylizedNeRF}: Consistent {3D} scene stylization as stylized {NeRF}
 via {2D}-{3D} mutual learning},
  author={Huang, Yi-Hua and He, Yue and Yuan, Yu-Jie and Lai, Yu-Kun and Gao, Lin},
  booktitle={CVPR},
  year={2022}
}

@inproceedings{t4,
  title={{PaletteNeRF}: Palette-based Color Editing for {NeRFs}},
  author={Wu, Qiling and Tan, Jianchao and Xu, Kun},
  booktitle={CVPR},
  year={2023}
}

@inproceedings{sine,
  title={{SINE}: Semantic-driven image-based {NeRF} editing with prior-guided editing field},
  author={Bao, Chong and Zhang, Yinda and Yang, Bangbang and Fan, Tianxing and Yang, Zesong and Bao, Hujun and Zhang, Guofeng and Cui, Zhaopeng},
  booktitle={CVPR},
  year={2023}
}

@inproceedings{arf,
  title={{ARF}: Artistic radiance fields},
  author={Zhang, Kai and Kolkin, Nick and Bi, Sai and Luan, Fujun and Xu, Zexiang and Shechtman, Eli and Snavely, Noah},
  booktitle={ECCV},
  year={2022}
}

@inproceedings{snerf,
  title={{SNeRF}: Stylized neural implicit representations for {3D} scenes},
  author={Nguyen-Phuoc, Thu and Liu, Feng and Xiao, Lei},
  booktitle={WACV},
  year={2022}
}

@article{shop,
  title={{NeRFshop}: Interactive Editing of Neural Radiance Fields},
  author={Jambon, Cl{\'e}ment and Kerbl, Bernhard and Kopanas, Georgios and Diolatzis, Stavros and Leimk{\"u}hler, Thomas and Drettakis, George},
  journal={Proceedings of the ACM on Computer Graphics and Interactive Techniques},
  year={2023}
}

@inproceedings{drag1,
  title={DragD3D: Vertex-based Editing for Realistic Mesh Deformations using 2D Diffusion Priors},
  author={Xie, Tianhao and Belilovsky, Eugene and Mudur, Sudhir and Popa, Tiberiu},
  booktitle={arXiv preprint arXiv:2310.04561},
  year={2023}
}

@inproceedings{pds,
    title={Posterior Distillation Sampling},
    author={Koo, Juil and Park, Chanho and Sung, Minhyuk},
    year={2024},
    booktitle={CVPR},
}

@inproceedings{mip,
    title={Mip-NeRF 360: Unbounded Anti-Aliased Neural Radiance Fields},
    author={Jonathan T. Barron and Ben Mildenhall and 
            Dor Verbin and Pratul P. Srinivasan and Peter Hedman},
    booktitle={CVPR},
    year={2022}
}

@article{tank,
    author    = {Arno Knapitsch and Jaesik Park and Qian-Yi Zhou and Vladlen Koltun},
    title     = {Tanks and Temples: Benchmarking Large-Scale Scene Reconstruction},
    journal   = {ACM Transactions on Graphics},
    year      = {2017},
}
\bibliographystyle{iclr2025_conference}

\appendix
\newpage

\section{Demo Video}
\label{sec:a}
A \textbf{demo video} of our framework description and editing results is included in the supplementary material.
\section{Additional Experiments}

\subsection{Qualitative Results on Large Object Movements}
\begin{figure}[ht]
\begin{center}
\includegraphics[width=.8\textwidth]{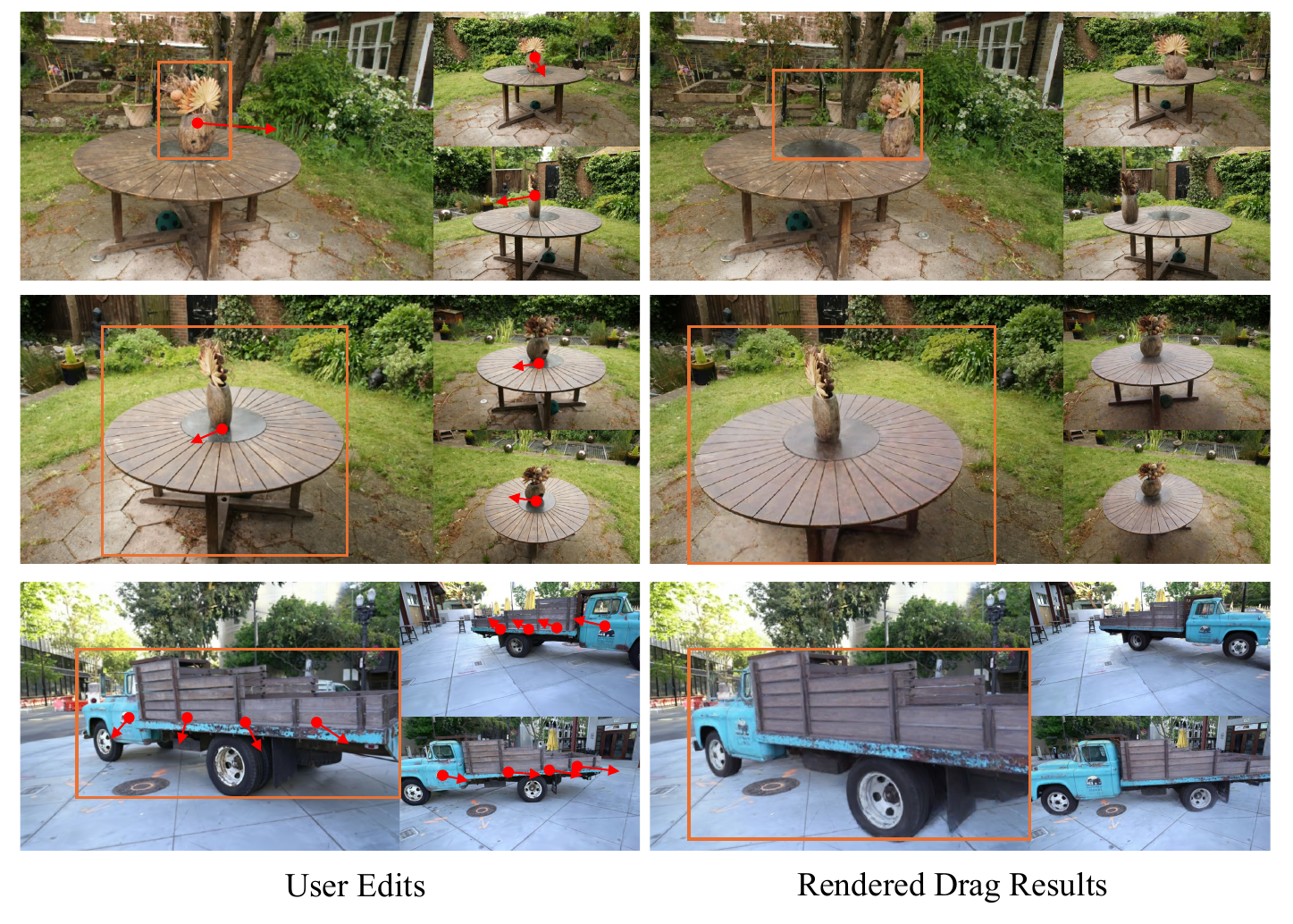}
\end{center} 
\vspace{-5mm}
\caption{\rebuttal{\textbf{Additional qualitative results for larger movements and larger objects:} Our method succeeds in longer-range movements like moving the flowerpot and large object movements like moving the table.}}
\label{fig:reut_large}
\end{figure}

\rebuttal{
We further conduct experiments on object movements, specifically including large movements of both regular-sized and large objects. As shown in Figure~\ref{fig:reut_large}, our method succeeds in handling long-distance movements, such as repositioning the flowerpot. Furthermore, for very large objects like the truck and the table,  our approach effectively moves them in a specified direction while minimizing artifacts in the removed and placed regions. These results showcase the generalizability of our method.
}
\subsection{Ablation on Dataset Editing Strategy}
\begin{figure}[ht]
\centering
\vspace{-2mm}
\includegraphics[width=.8\textwidth]{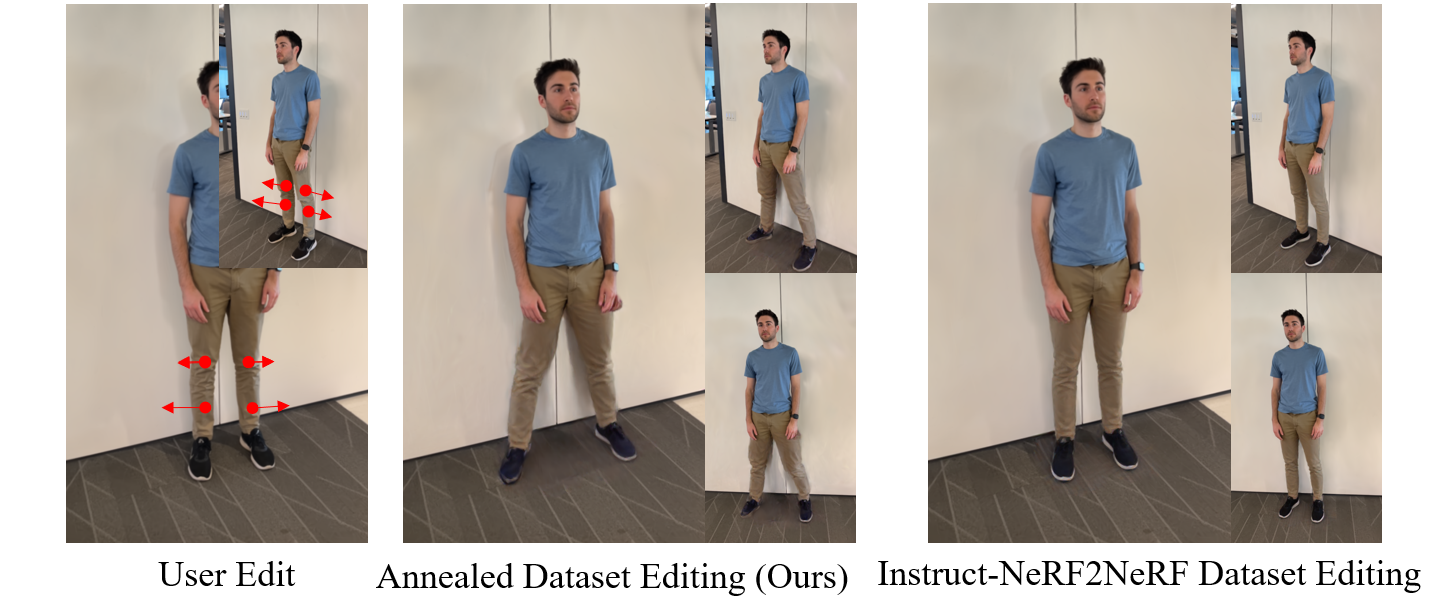}

\vspace{-3mm}
\caption{\rebuttal{\textbf{Ablation on dataset editing strategies:} Iterative dataset editing from Instruct-NeRF2NeRF~\citep{in2n} leads to degenerated results. In contrast, our annealed dataset editing maintains the geometry change.}}

\label{fig:rebut_anneal}
\end{figure}

\rebuttal{
To validate the importance of our dataset editing strategy, we conduct a comparison between using our annealed dataset editing and the iterative dataset editing~\citep {in2n}. As shown in Figure~\ref{fig:rebut_anneal}, editing one frame each time cannot change the geometry due to inconsistent constraints from other unedited views, leading to a degenerated result in the original scene. Instead, our method can successfully make the edits.
}
\subsection{Quantitative Ablation on Local Editing}
\rebuttal{
To further demonstrate the effectiveness of our local editing, we conduct quantitative evaluation on the ``extend sleeves'' edit. Specifically, we calculate the similarity between the rendered edited result and the originally rendered image in the unedited pixels. As shown in Table~\ref{tab:local_edit}, our local editing strategy demonstrates a strong capability to preserve the unedited regions and backgrounds effectively. 
}
\begin{table}[ht]
\centering

\begin{tabular}{l|ccc}
\toprule
& \textit{SSIM}$\uparrow$ &\textit{PSNR}$\uparrow$ &\textit{LPIPS}$\downarrow$\\ \midrule
Local Editing & \textbf{0.995} &\textbf{43.43} &\textbf{0.004}\\
Non-Local Editing &0.901 &24.44 &0.158 \\

\bottomrule
\end{tabular}
\caption{\rebuttal{\textbf{Quantitative ablation on local editing}: Our local editing strategy demonstrates a strong capability to preserve the unedited regions and backgrounds effectively.}}
\vspace{-3mm}
\label{tab:local_edit}
\end{table}

\subsection{Scaled User Study}
\begin{figure}[ht]
\centering

\includegraphics[width=.7\textwidth]{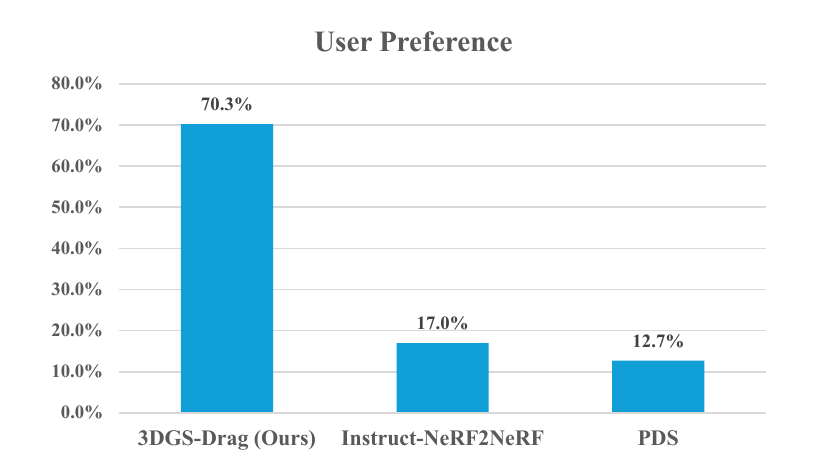}

\vspace{-3mm}
\caption{\rebuttal{\textbf{Scaled user study with 99 participants:} Our method still achieves significantly better preference over the baselines.}}

\label{fig:scaled_user}
\end{figure}

\rebuttal{
To improve the generalizability of our user study, we increased the number of participants from 19 to 99. As shown in Figure~\ref{fig:scaled_user}, the key conclusion that our method surpasses previous baselines remains the same.
}

\subsection{Comparison with 2D Drag Methods}\label{sec:sup_drag}
\begin{figure}[ht]
\begin{center}
\includegraphics[width=.75\textwidth]{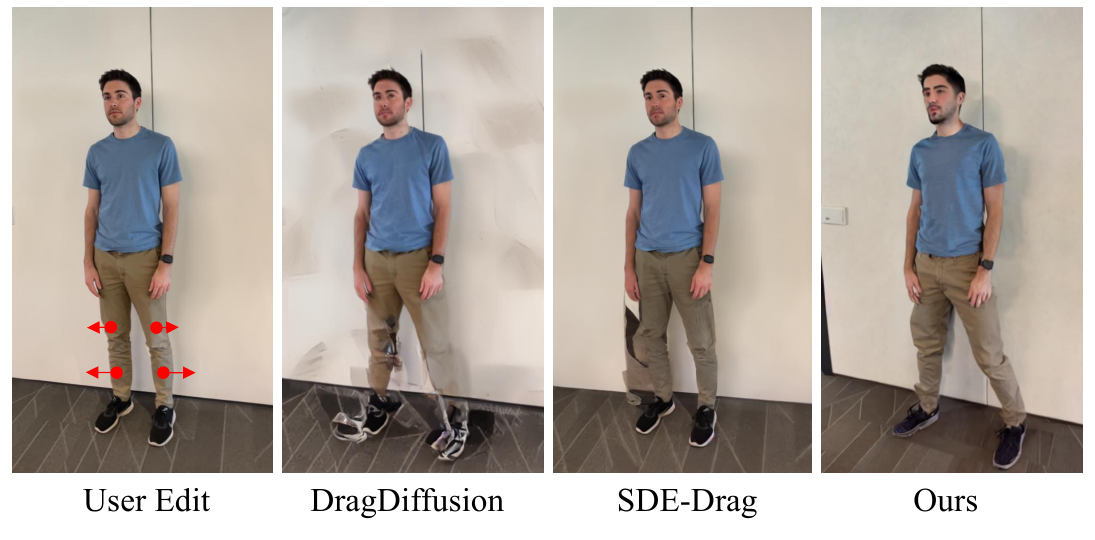}
\end{center} 
\vspace{-5mm}
\caption{\textbf{Comparison on 2D drag results:} Compared with recent 2D drag methods, our method does not need a time-consuming inverse-forward process and produces more consistent results. In comparison, the baseline DragDiffusion generates noisy legs and floor. SDE-Drag succeeds in maintaining the background but inserts objects in the hand and does not correctly move the leg.}
\vspace{-10pt}
\label{fig:2D_drag}
\end{figure}

We conduct qualitative comparisons on 2D drag edits to verify our method's effectiveness. Specifically, we focus on our deformation-guided diffusion editing quality, comparing the result with the previous 2D drag methods, Dragdiffusion~\citep{dragdif} and SDE-Drag~\citep{sde}. As shown in Figure~\ref{fig:2D_drag}, DragDiffusion tends to create many more artifacts and unrelated textures, while our result is cleaner and matches better with the edit. Considering SDE-Drag~\citep{sde}, it fails to move the leg correctly and instead generates an object at the location. The results show that our method can achieve better consistency by directly operating at the image level. 

\begin{figure}[ht]
\centering
\includegraphics[width=0.8\textwidth]{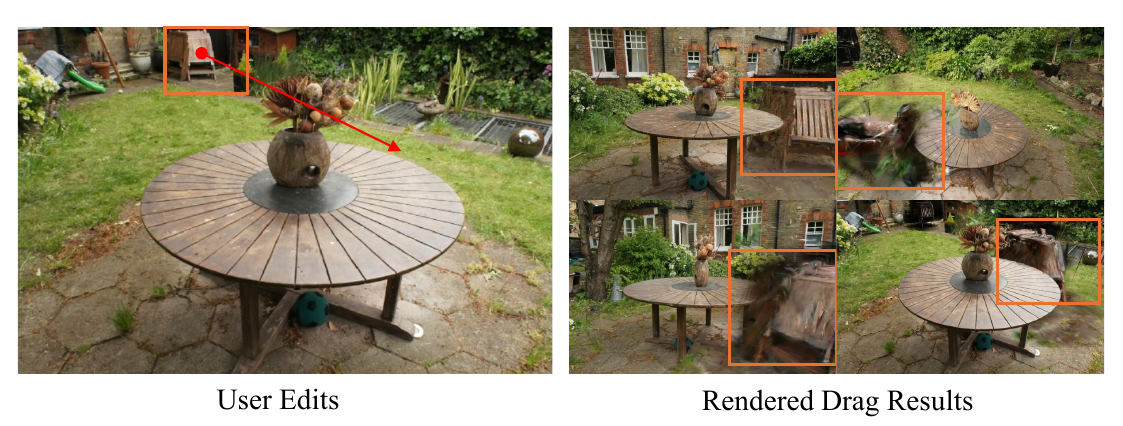}

\vspace{-1mm}
\caption{\rebuttal{\textbf{Limitation on generating unseen side:} When dragging background objects with an unseen side to the foreground, the results from other views are incorrect.}}

\label{fig:sup_limi_1}
\end{figure}
\begin{figure}[ht]
\centering
\includegraphics[width=0.8\textwidth]{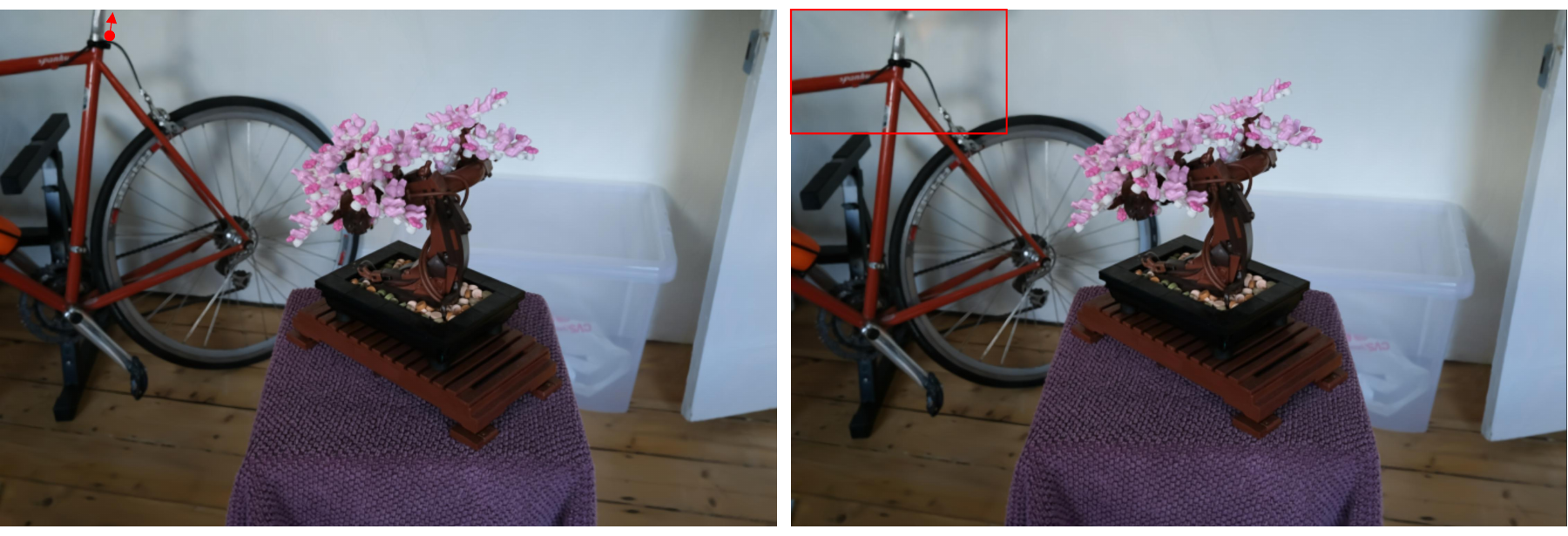}

\vspace{-1mm}
\caption{\rebuttal{\textbf{Limitation on dragging objects outside the border:} Refining and optimizing become challenging when dragging towards the unseen or border area.}}

\label{fig:sup_limi_2}
\end{figure}
\section{In-Depth Discussion of Limitations}
\label{sec:e}
\rebuttal{
Our method encounters two specific failure cases: generating the unseen side of an object and dragging objects outside the border area. Here, we present qualitative results to further elucidate these limitations. 
}
\subsection{Generating Unseen Side}
\rebuttal{
As shown in Figure~\ref{fig:sup_limi_1}, when moving the wooden support to the foreground, the unseen portions (e.g., the back of the object) are rendered incorrectly and exhibit noticeable artifacts. This occurs because there are no 3D Gassians to represent the unseen back side. Consequently, the deformed result contains meaningless patterns that cannot be corrected through the diffusion process.
}
\subsection{Dragging Objects Outside the Border}
\rebuttal{
As shown in Figure ~\ref{fig:sup_limi_2}, we struggle to refine the artifacts when only a portion of the object is visible at the border. This is because the diffusion model has less capability to correct the boundary part due to the ambiguity in interpolating the part to the whole. In addition, since it is observed by sparser views, it raises further challenges for optimizing 3D Gaussians.
}

\subsection{Potential Biases in Quantitative Evaluations}
\rebuttal{
While we employ both human preference scores and automated metrics from GPT evaluation, there could still be potential biases, such as those arising from the participant selection process or the specific version and training data of the GPT models used. This challenge is common in generative modeling, where there is no ground truth. Future work would benefit from conducting larger-scale human studies with more diverse participant pools and developing more comprehensive evaluation protocols that can better assess both geometric accuracy and visual fidelity of 3D edits.
}
\section{Additional Deformation Details}\label{sec:sup_deform}
The calculation of relative rotation $\Delta q_h^{ik}$ is briefly described as follows. Given handle points $p_h^i$ and $p_h^k$ and their corresponding target points $p_t^i$ and $p_t^k$ respectively, we firstly calculate the unit vectors as
\begin{equation}
v_h^{ik} = \frac{p_h^k - p_h^i}{\|p_h^k - p_h^i\|}
\quad \text{and} \quad
v_t^{ik} = \frac{p_t^k - p_t^i}{\|p_t^k - p_t^i\|}
\end{equation}
Next, we compute the cross product and the dot product of these two unit vectors:
\begin{equation}
\mathbf{r} = v_h^{ik} \times v_t^{ik},\quad 
s = v_h^{ik} \cdot v_t^{ik} 
\end{equation}
We then construct the quaternion \(\Delta q_h^{ik}\) by combining the dot product and cross product:

\begin{equation}
\Delta q_h^{ik} = [s, \mathbf{r}_x, \mathbf{r}_y, \mathbf{r}_z]
\end{equation}

Finally, we standardize and normalize the quaternion to ensure it has unit length.

\section{Social Impact and Future Work}
\noindent\textbf{Future Work.} For future work, we plan to extend current progressive editing capabilities to generate 3D animations. As \ours is able to move or modify objects progressively, it is possible to generate long-term trajectories and human motions. In addition, we will focus on improving the model's scalability to accommodate larger scenes with dynamic objects and shadow effects. 

\noindent\textbf{Potential Social Impact.} The potential societal impact of \ours spans across multiple dimensions. Designed as a fine-grained editing model, our \ours offers convenient manipulation of 3D scenes and robust support for AR applications. In addition, given the rapid development and widespread adoption of 3D Gaussians, our method seamlessly integrates with this ecosystem. With its user-friendly interface requiring only the selection of handle and target points, our model is accessible even to untrained individuals.

\end{document}